\documentclass[11pt]{article}
\usepackage{acl2016}
\usepackage{times}
\usepackage{url}
\usepackage{latexsym}
\usepackage{graphicx}
\usepackage{amsmath}
\usepackage{amsfonts}
\usepackage{comment}
\usepackage{multirow}
\usepackage{array}
\aclfinalcopy 
\def\aclpaperid{773} 

\title{Recurrent neural network models for disease name recognition using domain invariant features}

\author{Sunil Kumar Sahu \and Ashish Anand \\
  Department of Computer Science and Engineering\\
  Indian Institute of Technology Guwahati\\
  Assam, India  - 781039\\
  {\tt \{sunil.sahu, anand.ashish\}@iitg.ernet.in }}

\date{}

\begin{document}
\maketitle---
\begin{abstract}
 Hand-crafted features based on linguistic and domain-knowledge play crucial role in determining the performance of disease name recognition systems. Such methods are further limited by the scope of these features or in other words, their ability to cover the contexts or word dependencies within a sentence. In this work, we focus on reducing such dependencies and propose a domain-invariant framework for the disease name recognition task. In particular, we propose various end-to-end recurrent neural network (RNN) models for the tasks of disease name recognition and their classification into four pre-defined categories. We also utilize convolution neural network (CNN) in cascade of RNN to get character-based embedded features and employ it with word-embedded features in our model. We compare our models with the state-of-the-art results for the two tasks on NCBI disease dataset. Our results for the disease mention recognition task indicate that state-of-the-art performance can be obtained without relying on feature engineering. Further the proposed models obtained improved performance on the classification task of disease names.  
\end{abstract}

\section{Introduction}
Automatic recognition of disease names in biomedical and clinical texts is of utmost importance for development of more sophisticated NLP systems such as information extraction, question answering, text summarization and so on \cite{Rosario04}. Complicate and inconsistent terminologies, ambiguities caused by use of abbreviations and acronyms, new disease names, multiple names (possibly of varying number of words) for the same disease, complicated syntactic structure referring to multiple related names or entities are some of the major reasons for making automatic identification of the task difficult and challenging \cite{leaman09}. State-of-the-art disease name recognition systems \cite{MahbubChowdhury10,Dogan12,Dogan14} depends on user defined features which in turn try to capture context keeping in mind above mentioned challenges. Feature engineering not only requires linguistic as well as domain insight but also is time consuming and is corpus dependent.

Recently window based neural network approach of 
\cite{collobert08,collobert11a} got lot of attention in different sequence tagging tasks in NLP. It gave state-of-art results in many sequence labeling problems without using many hand designed or manually engineered features. One major drawback of this approach is its inability to capture features from outside window. Consider a sentence {\it ``Given that the skin of these adult mice also exhibits signs of de novo hair-follicle morphogenesis, we wondered whether human pilomatricomas might originate from hair matrix cells and whether they might possess beta-catenin-stabilizing mutations''} (taken {\em verbatim} from PMID: 10192393), words such as {\it signs} and {\it originate} appearing both sides of the word {\it``pilomatricomas"}, play important role in deciding it is a disease. Any model relying on features defined based on words occurring within a fixed window of neighboring words will fail to capture information of influential words occurring outside this window.

Our motivation can be summarized in the following question: {\em can we identify disease name and categorize them without relying on feature engineering, domain-knowledge or task specific resources}? In other words, we can say this work is motivated towards mitigating the two issues: first, feature engineering relying on linguistic and domain-specific knowledge; and second, bring flexibility in capturing influential words affecting model decisions irrespective of their occurrence anywhere within the sentence. For the first, we used character-based embedding (likely to capture orthographic and morphological features) as well as word embedding (likely to capture lexico-semantic features) as features of the neural network models. 

For the second issue, we explore various recurrent neural network (RNN) architectures for their ability to capture long distance contexts. We experiment with bidirectional RNN (Bi-RNN), bidirectional long short term memory network (Bi-LSTM) and bidirectional gated recurrent unit (Bi-GRU). In each of these models we used sentence level log likelihood approach at the top layer of the neural architecture. The main contributions of the work can be summarized as follows
\begin{itemize}
  \item Domain invariant features with various RNN architectures for the disease name recognition and classification tasks,
  \item Comparative study on the use of character based embedded features, word embedding features and combined features in the RNN models.
  \item Failure analysis to check where exactly our models are failed in the considered tasks.
\end{itemize}
Although there are some related works (discussed in sec~\ref{sec:rel_work}), this is the first study, to the best of our knowledge, which comprehensively uses various RNN architectures without resorting to feature engineering for disease name recognition and classification tasks. 

Our results show near state-of-the-art performance can be achieved on the
disease name recognition task. More significantly, the proposed models obtain
significantly improved performance on the disease name classification task.

\section{Methods}
We first give overview of the complete model used for the two tasks. Next we explained embedded features used in different neural network models. We provide short description of different RNN models in the section~\ref{sec:nnm}. Training and inference strategies are explained in the section~\ref{sec:train}. 

\subsection{Model Architectures}
Similar to any named entity recognition task, we formulate the disease mention recognition task as a token level sequence tagging problem. Each word has to be labeled with one of the defined tags. We choose {\it BIO model} of tagging, where {\it B} stands for beginning, {\it I} for intermediate and {\it O} for outsider or other. This way we have two possible tags for all entities of interest, i.e., for all disease mentions, and one tag for other entities.

\begin{figure}[!htb] 
\begin{center}
\includegraphics[width=0.5\textwidth]{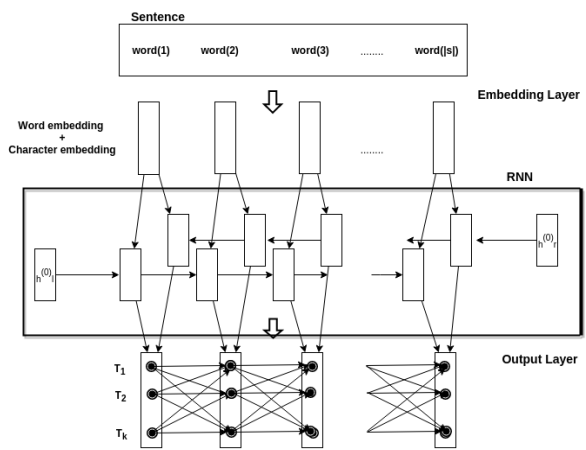}
\caption{Generic bidirectional recurrent neural network with sentence level log likelihood at the top-layer for sequence tagging task}
\label{fig:gen-arch}
\end{center}
\end{figure}

Generic neural architecture is shown in the figure~\ref{fig:gen-arch}. In the very first step, each word is mapped to its embedded features.

We call this layer as {\it embedding layer}. This layer acts as input to hidden layers 
of RNN model. We study the three different RNN models, and have described them briefly
in the section~\ref{sec:nnm}. Output of the hidden layers is then fed to the output layer to
compute the scores for all tags of interest~\cite{collobert11a,HuangXY15}. In output 
layer we are using sentence level log likelihood, to make inference.
Table~\ref{tab:my-label} briefly describes all notations used in the paper.

\begin{table}[ht]
\centering
\begin{tabular}{|c|m{4.2cm}|}
\hline
Symbols & Explanation \\ \hline
V & Vocabulary of words ${\it ( v_1,v_2...v_{|V|})}$   \\ \hline
C & Vocabulary of characters ${( c_1, c_2..c_{|C|} )}$ \\ \hline
T & Tag set ${(t_1, t_2...t_{|T|})}$ \\ \hline
$d^{we}$ & Dimension of word embedding \\ \hline
$d^{chr}$ & Dimension of character embedding \\ \hline
$d^{ce}$ & Dimension of character level word embedding \\ \hline
$M^{we} \in \mathbb{R}^{d^{we}X|V|}$ &  word embedding matrix, where every column $M^{we}_i$ is a vector representation of corresponding word ${\it v_i}$ in {\it V} \\ \hline
$M^{cw} \in \mathbb{R}^{d^{chr}X|C|}$ & character embedding matrix, where every column $M^{cw}_i$ is a vector representation of corresponding character ${\it c_i}$ in {\it C}.\\ \hline
$w^{(i)} \in \mathbb{R}^{d^{we}}$ &  word embedding of $v_i$ \\ \hline
$y^{(i)} \in \mathbb{R}^{d^{ce}}$ & character level word embedding of $v_i$ \\ \hline
$x^{(i)} \in \mathbb{R}^{d^{we}+d^{ce}}$ & feature vector of word $w^{(i)}$. We get this after concatenating $w^{(i)}$ and $y^{(i)}$ \\ \hline
$z^{(i)} \in \mathbb{R}^{|T|}$ & score for $i^{th}$ word in sentence at output layer of neural network. Here ${\it j}^{th}$ element will indicate the score for $t_j^{th}$ tag. \\ \hline
$W^{*}_{*},U^{*}_{*},V^{*}_{*}$ & Parameters of different neural networks \\ \hline
\end{tabular}
\caption{Notation}
\label{tab:my-label}
\end{table}







\subsection{Features}
\subsection*{Distributed Word Representation (WE)}
Distributed word representation or word embedding or simply word vector \cite{Bengio03,collobert08} is the technique of learning vector representation of a word in a given corpus. 
Word vectors are present in columns of matrix $M^{we}$. We can get this vector by taking product of matrix $M^{we}$ and {\it one hot vector} 
of $v_i$.
\begin{equation}
w^{(i)} = M^{we}\,h^{(i)}
\end{equation}
Here $h^{(i)}$ is the one hot vector representation of $i^{th}$ word in {\it V}. We use pre-trained $50$ dimensional word vectors learned using skipgram method on a biomedical corpus \cite{mikolov13a,mikolov13b,muneeb15}.  

\subsection*{Character Level Word Embedding (CE)}
Word embedding preserve syntactic and semantic information well but fails to seize morphological and shape information. However, for the disease entity recognition task, such information can play an important role.  For instance, letter {\it -o-} in the word {\em gastroenteritis} is used to combine various body parts {\it gastro} for 
{\it stomach}, {\it enter} for {\it intestines}, and {\em itis} indicates {\em inflammation}. Hence taken together it implies {\em inflammation of stomach and intestines}, where {\em -itis} play significant role in determining it is actually a disease name.

\begin{figure}[!htb] 
\begin{center}
\includegraphics[width=0.5\textwidth]{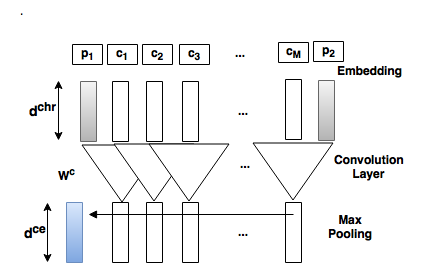}
\caption{CNN with Max Pooling for Character Level Embedding ($p_1$ and $p_2$ are padding). Here, filter length is 3.}
\label{fig:char-embed}
\end{center}
\end{figure}

Character level word embedding was first introduced by \cite{dos2014} with the motivation to capture word shape and morphological features in word embedding. Character level word embedding also automatically mitigate the problem of out of vocabulary words as we can embed any word by its characters through character level embedding. In this case, a vector is initialized for every character in the corpus. Then we learn vector representation for any word by applying CNN on each vector of character sequence of that word as shown in figure \ref{fig:char-embed}. These character vectors will get update while training RNN in supervised manner only. Since number of characters in the dataset is not high we assume that every character vectors will get sufficient updation while training RNN itself.

Let $\{p_1, c_1, c_2 ... c_M, p_2\}$ is sequence of characters for a word with padding at beginning and ending of word and let $\{a_l, a_1, a_2...a_M, a_r\}$ is its sequence of character vector, which we obtain by multiplying $M^{cw}$ with one hot vector of corresponding character. To obtain character level word embedding we need to feed this in convolution neural network (CNN) with max pooling layer~\cite{dos2014}. Let $W^c \in \mathbb{R}^{d^{ce}X(d^{chr}Xk)} $ is a filter and $b^c$ bias of CNN, then 
\begin{equation}
[y^{(i)}]_j = \max_{1<m<M} [W^c q^{(m)} + b^c]_j
\end{equation}
Here $k$ is window size, $q^{(m)}$ is obtained by concatenating the vector of $(k-1)/2$ character left to $(k-1)/2$ character right of $c_m$. Same filter will be used for all window of characters and max pooling operation is performed on results of all. We learn $100$ dimensional character embedding for all characters in a given dataset (avoiding case sensitivity) and $25$ dimensional character level word embedding from character sequence of words.

\subsection{Recurrent Neural Network Models}
\label{sec:nnm}
Recurrent Neural Network (RNN) is a class of artificial neural networks which utilizes 
sequential information and maintains history through its intermediate layers 
\cite{Graves:2009,Graves13}. We experiment with three different variants of RNN,
which are briefly described in subsequent subsections.
 
\subsection*{Bi-directional Recurrent Neural Network}
In Bi-RNN, context of the word is captured through past and future words. This is achieved by having two hidden components in the intermediate layer, as schematically shown in the fig~\ref{fig:gen-arch}. One component process the information in forward direction (left to right) and other in reverse direction. Subsequently outputs of these components then concatenated and fed to the output layer to get score for all tags of the considered word. Let $x^{(t)}$ is a feature vector of $t^{th}$ word in sentence (concatenation of corresponding embedding features $w^{t_{i}}$ and $y^{t_{i}}$) and $h^{(t-1)}_l$ is the computation of last hidden state at $(t-1)^{th}$ word, then computation of hidden and output layer values would be:
\begin{align*} 
h_{l}^{(t)} = tanh(U^{l}x^{(t)} + W^{l}h_{l}^{(t-1)})
\end{align*}
\begin{equation} 
\label{rnn_score}
z^{(t)} = V(h_{l}^{(t)}:h_{r}^{(t)})
\end{equation}
Here $U^l \in \mathbb{R}^{n_H \times n_I}$ and $W^l \in \mathbb{R}^{n_H \times n_H}$, where $n_I$ is input vector of length $d^{we}+d^{ce}$, $n_H$ is hidden layer size and $V \in \mathbb{R}^{n_O \times (n_H+n_H)}$ is the output layer parameter. $h_{l}^{(t)}$ and $h_{r}^{(t)}$ correspond to left and right hidden layer components respectively and $h_{r}^{(t)}$ is calculated similarly to $h_{l}^{(t)}$ by reversing the words in the sentence. At the beginning $h^{(0)}_l$ and $h^{(0)}_r$ are initialized randomly.

\subsection*{Bi-directional Long Short Term Memory Network}
Traditional RNN models suffer from both vanishing and exploding gradient \cite{bengio2013,bengio2013advances}. Such models are likely to fail where we need longer contexts to do the job. These issues were the main motivation behind the LSTM model \cite{Hochreiter97}. LSTM layer is  just another way to compute a hidden state which introduces a new structure called a memory cell ($c_t$) and three gates called as input ($i_t$), output ($o_t$) and forget ($f_t$) gates. 
 
These gates are composed of sigmoid activation function and responsible for regulating information in memory cell. The input gate by allowing incoming signal to alter the state of the memory cell, regulates proportion of history information memory cell will keep. On the other hand, the output gate regulates what proportion of stored information in the memory cell will influence other neurons. Finally, the forget gate can modulate the memory cell’s and allowing the cell to remember or forget its previous state. Computation of memory cell ($c^{(t)}$) is done through previous memory cell and candidate hidden state ($g^{(t)}$) which we compute through current input and the previous hidden state. The final output of hidden state would be calculated based on memory cell and forget gate.
 
In our experiment we used model discussed in \cite{Graves13,HuangXY15}.
Let $x^{(t)}$ is feature vector for $t^{th}$ word in a sentence and $h_l^{(t-1)}$ is previous hidden state then computation of hidden ($h_l^{(t)}$) and output layer ($z^{(t)}$) of LSTM would be. 
\begin{align*} 
&i^{(t)}_l = \sigma ( U^{(i)}_l x^{(t)} +  W^{(i)}_l h_l^{(t-1)} + b_l^i)\\
&f^{(t)}_l = \sigma (U^{(f)}_l x^{(t)} + W^{(f)}_l h_l^{(t-1)} + b_l^f)\\
&o^{(t)}_l = \sigma (U^{(o)}_l x^{(t)}  + W^{(o)}_l h_l^{(t-1)} + b_l^o)\\
&g^{(t)}_l = tanh(U_l^{(g)} x^{(t)} +  W^{(g)}_l h_l^{(t-1)} + b_l^{g}) \\
&c^{(t)}_l = c^{(t-1)}_l * f_l + g_l * i_l \\
&h^{(t)}_l = tanh(c^{(t)}_l) * o_l 
\end{align*} 
Where $\sigma $ is sigmoid activation function, $*$ is a element wise product, $U^{(i)}_l, U^{(f)}_l, U^{(o)}_l, U^{(g)}_l \in \mathbb{R}^{n_H \times n_I}$ and $W^{(i)}_l, W^{(o)}_l, W^{(f)}_l, W^{(g)}_l \in \mathbb{R}^{n_H \times n_H}$, where $n_I$ is input size ($d^{we}+d^{ce}$) and $n_H$ is hidden layer size. We compute $h_{r}^{(t)}$ in similar manner as $h_l^{(t)}$ by reversing the all words of sentence. Let $V \in \mathbb{R}^{n_O \times (n_H+n_H)}$ ($n_O$ size of output layer) is the parameter of output layer of LSTM then computation of output layer will be: 
\begin{equation} 
\label{lstm_score}
z^{(t)} = V(h_{l}^{(t)}:h_{r}^{(t)})
\end{equation}

\subsection*{Bi-directional Gated Recurrent Unit Network}
A gated recurrent unit (GRU) was proposed by \cite{ChoMBB14} to make each recurrent unit to adaptively capture dependencies of different time scales. Similar to the LSTM unit, the GRU has gating units reset {\it r} and update {\it z} gates
that modulate the flow of information inside the unit, however, without having a separate memory cells. The resulting model is simpler than standard LSTM models.

We follow \cite{ChungGCB14} model of GRU to transform the extracted word embedding and character embedding features to score for all tags. 
Let $x^{(t)}$ embedding feature for {\it t}th word in sentence and $h_l^{(t-1)}$ is computation of hidden state for $(t-1)$th word then computation of GRU would be:

\begin{align*}
&z^{(t)}_l = \sigma ( U^{(z)}_l x^{(t)}  +  W^{(z)}_l h_l^{(t-1)} + b_l^{(z)} )\\
&r^{(t)}_l = \sigma ( U^{(r)}_l x^{(t)}  +  W^{(r)}_l h_l^{(t-1)} + b_l^{(r)} )\\
&\tilde{h}^{(t)}_l = \tanh ( U^{(h)}_l x^{(t)}  + W^{(h)}_{l} h_l^{(t-1)} * r_l  + b_l^{(h)}) \\ 
&h^{(t)}_l = z^{(t)}_l * \tilde{h}_l + (1-z^{(t)}_l) * h_l^{(t-1)} 
\end{align*}

\begin{equation}
\label{gru_score}
z^{(t)} = V(h_{l}^{(t)}:h_{r}^{(t)})
\end{equation} 
Where $*$ is pair wise multiplication, $U^{(z)}_l, U^{(r)}_l,$ $ U^{(h)}_l, U^{(h)}_l$ $\in \mathbb{R}^{n_H \times n_I}$ and $W^{(z)}_l, W^{(r)}_l W^{(h)}_l$ $\in \mathbb{R}^{n_H \times n_H}$ are parameters of GRU. $V \in \mathbb{R}^{n_O \times (n_H+n_H)}$ is output layer parameter.
Computation of $h_{r}^{(t)}$ is done in similar manner as $h_{l}^{(t)}$ by reversing the words of sentence.

\subsection{Training and Inference}
\label{sec:train}
Equations~\ref{rnn_score}, \ref{lstm_score} and \ref{gru_score} are the scores of all possible tags for $t^{th}$ word sentence. We follow {\it sentence-level log-likelihood (SLL)} \cite{collobert11a} approach equivalent to linear-chain CRF to infer the scores of a particular tag sequence for the given word sequence. Let $[w]_{1}^{|s|}$ is sentence and $[t]_{1}^{|s|}$ is the tag sequence for which we want to find the joint score, then score for the whole sentence with the particular tag sequence would be: 
\begin{equation}
s( [w]_{1}^{|s|} , [t]_{1}^{|s|}) = \sum_{1 \le i \le |s|} (W_{t_{i-1},t_{i}}^{trans} + z^{(i)}_{t_i} ),
\end{equation} 
where $W^{trans}$ is transition score matrix and $W^{trans}_{i,j}$ is indicating the transition score moving from tag $t_i$ to $t_j$; $t_j$ is tag for the $j^{th}$ word; $z^{(i)}_{t_i}$ is the output score from the neural network model for the tag $t_i$ of $i^{th}$ word. 
To train our model we used cross entropy loss function and adagrad \cite{Duchi10} approach to optimize the loss function. Entire neural network parameters, word embedding, character embedding and $W^{trans}$ (transition score matrix used in the SLL) was updated during training. Entire code has been implemented using theano \cite{Bastien-Theano-2012} library in python language.

\section{Experiments}
\subsection{Dataset}
We used NCBI dataset \cite{Dogan12}, the most comprehensive publicly available dataset annotated with disease mentions, in this work. NCBI dataset has been manually annotated by a group of medical practitioners for identifying diseases and their types in biomedical articles. All disease mentions were categorized into four different categories, namely, {\it specific disease}, {\it disease class}, {\it composite disease} and {\it modifier}. A word is annotated as {\it specific disease}, if it indicates a particular disease. {\it Disease class} category indicates a word describing a family
of many specific diseases, such as autoimmune disorder. A string signifying two or more different disease mentions is annotated with {\it composite mention}. {\it Modifier} category indicates disease mention has been used as modifiers for other concepts. This dataset is a extension of the AZDC dataset \cite{leaman09} which was annotated with  disease mentions only and not with their categories. Statistics of the dataset is mentioned in the Table \ref{tab:tab1}.

\begin{table}[ht]
\begin{tabular}{|c|c|c|c|c|}
\hline
\textit{\textbf{Corpus}} & \textit{\textbf{Train set}} & \textit{\textbf{Dev set}} & \textit{\textbf{Test set}} \\ \hline
sentences  		& 5661 & 939 & 961 \\ \hline
disease 		& 5148 & 791 & 961 \\ \hline
spe. dis.		& 2959 & 409 & 556 \\ \hline
disease class 	& 781  & 127 & 121 \\ \hline
modifier		& 1292 & 218 & 264 \\ \hline
comp. men. 		& 116  & 37  & 20  \\ \hline
\end{tabular}
\caption{Dataset statistics. spe. dis. : specific disease and comp. men.: composite mention }
\label{tab:tab1}
\end{table}

\begin{table*}[t]
\begin{minipage}{\textwidth}
\centering
\begin{tabular} 
{|p{0.04\linewidth}|p{0.14\linewidth}|p{0.09\linewidth}|p{0.09\linewidth}|p{0.09\linewidth}|p{0.09\linewidth}|p{0.09\linewidth}|p{0.09\linewidth}|}\hline
\multirow{2}{*}{\textbf{Task}} & \multirow{2}{*}{\textbf{Model}} & \multicolumn{3}{c|}{\textbf{Validation Set}} & \multicolumn{3}{c|}{\textbf{Test Set}} \\ \cline{3-8} 
& & \textbf{Precision} & \textbf{Recall} & \textbf{F1 Score} & \textbf{Precision} & \textbf{Recall} & \textbf{F1 Score} \\ \hline 
\hline
\multirow{4}{*}{\textbf{A}} & NN+CE & 76.98 & 75.80 & 76.39 & 78.51 & 72.75 & 75.52 \\ \cline{2-8} 
& Bi-RNN+CE & 71.96 & 74.90 & 73.40 & 74.14 & 72.12 & 73.11\\ \cline{2-8} 
& Bi-GRU+CE & 76.28 & 74.14 & 75.19 & 76.03 & 69.81 & 72.79\\ \cline{2-8} 
& Bi-LSTM+CE& 81.52 & 72.86 & {\bf 76.94} & 76.98 & 75.80 & {\bf 76.39}   \\ \cline{2-8} 
\hline
\multirow{4}{*}{\textbf{B}} & NN+CE & 67.27 & 53.45 & 59.57 & 67.90 & 49.95 & 57.56 \\ \cline{2-8} 
& Bi-RNN+CE & 61.34 & 56.32 & 58.72 & 60.32 & 57.28 & 58.76 \\ \cline{2-8} 
& Bi-GRU+CE & 61.94 & 59.11 & {\bf 60.49} & 62.56 & 56.50 & 59.38 \\ \cline{2-8} 
& Bi-LSTM+CE& 61.82 & 57.03 & 59.33 & 64.74 & 55.53 & {\bf 59.78} \\ \hline
\end{tabular}
\caption{Performance of various models using 25 dimensional CE features, A:Disease name recognition, B: Disease classification task}
\label{tab:tab2}
\end{minipage}
\end{table*}
In our evaluation we used this dataset in two settings, {\bf A}: {\it disease mention recognition}, where all disease types are flattened into a single category and, the {\bf B}: {\it disease class recognition}, where we need to decide exact categories of disease mentions. It is noteworthy to mention that the Task B is more challenging as it requires model to capture semantic contexts to put disease mentions into appropriate categories.
 
 \section{Results and Discussion}
\subsection*{Evaluation of different models using CE}
We first evaluate the performance of different RNNs using only character embedding features. We compare the results of RNN models with window based neural network \cite{collobert11a} using sentence level log likelihood approach (NN + CE). For the window based neural network, we considered window size $5$ (two words from both left and right, and one central word) and same settings of character embedding were used as features. The same set of parameters are used in all experiments unless we mention specifically otherwise. We used exact matching scheme to evaluate performance of all models. 

Table \ref{tab:tab2} shows the results obtained by different RNN models with only character level word embedding features. For the task A (Disease name recognition) Bi-LSTM and NN models gave competitive performance on the test set, while Bi-RNN and Bi-GRU did not perform so well. On the other hand for the task B, there is $2.08\%-3.8\%$ improved performance (F1-score) shown by RNN models over the NN model again on the test set. Bi-LSTM model obtained F1-score of $59.78\%$ while NN model gave $57.56\%$. As discussed earlier, task B is difficult than task A as disease category is more likely to be influenced by the words falling outside the context window considered in window based methods. This could be reason for RNN models to perform well over the NN model. This hypothesis will be stronger if we observe similar pattern in our other experiments.
 
\begin{table*}[t]
\begin{minipage}{\textwidth}
\centering
\begin{tabular} 
{|p{0.04\linewidth}|p{0.2\linewidth}|p{0.09\linewidth}|p{0.08\linewidth}|p{0.09\linewidth}|p{0.09\linewidth}|p{0.07\linewidth}|p{0.10\linewidth}|}\hline
\multirow{2}{*}{\textbf{Task}} & \multirow{2}{*}{\textbf{Model}} & \multicolumn{3}{c|}{\textbf{Validation Set}} & \multicolumn{3}{c|}{\textbf{Test Set}} \\ \cline{3-8} 
& & \textbf{Precision} & \textbf{Recall} & \textbf{F1 Score} & \textbf{Precision} & \textbf{Recall} & \textbf{F1 Score} \\ \hline 
\hline
\multirow{4}{*}{\textbf{A}} & NN+WE & 81.86 & 76.82 & 79.26 & 80.32 & 73.58 & 76.81	\\\cline{2-8}
& Bi-RNN+WE & 84.14 & 77.46 & 80.67 & 82.49 & 73.58 & 77.78 \\ \cline{2-8}
& Bi-GRU+WE & 84.51 & 78.23 & 81.25 & 82.32 & 75.16 & 78.58 \\ \cline{2-8}
& Bi-LSTM+WE& 85.13 &77.72&{\bf 81.26}&84.87&74.11&{\bf 79.13}\\ \hline
\multirow{4}{*}{\textbf{B}} & NN+WE & 65.33 & 56.43 & 60.55 & 64.23 & 57.14 & 60.48 \\ \cline{2-8}
& Bi-RNN+WE	& 63.62 & 56.84 & 60.04 & 67.47 & 57.50 & 62.09 \\ \cline{2-8}
& Bi-GRU+WE	& 66.42 & 57.41 & 61.59 & 68.25 & 58.58 & 63.05 \\ \cline{2-8}
& Bi-LSTM+WE &  67.48 & 58.01 & {\bf 62.39} & 68.97 & 58.25 & {\bf 63.16} \\ \cline{2-8} 
\hline
\hline
\multirow{4}{*}{\textbf{A}} & NN+WE+CE 	& 76.37 & 78.62 & {\bf 77.48} & 74.92 & 75.16 & 75.04 \\ \cline{2-8}
& Bi-RNN+WE+CE 	& 76.10 & 75.03 & 75.56 & 77.01 & 72.33 & 74.59 \\ \cline{2-8}
& Bi-GRU+WE+CE 	& 77.73 & 76.44 & 77.08 & 78.04 & 73.38 & {\bf 75.63} \\\cline{2-8}
& Bi-LSTM+WE+CE & 76.94 & 77.34 & 77.14 & 76.10 & 74.11 & 75.09\\ 
 \hline
\multirow{4}{*}{\textbf{B}} & NN+WE+CE 		& 67.60 & 56.70 & 61.67 & 67.60 & 56.70 & 61.67 \\ \cline{2-8}
& Bi-RNN+WE+CE 	& 60.94 & 61.34 & 61.14 & 64.36 & 60.90 & 62.58 \\ \cline{2-8}
& Bi-GRU+WE+CE 	& 61.58 & 61.99 & {\bf 61.78} & 61.92 & 63.85 & {\bf 62.87} \\ \cline{2-8}

& Bi-LSTM+WE+CE & 64.92 & 58.61 & 61.60 & 61.14 & 60.54 & 60.84 \\
\hline
\end{tabular}
\caption{Performance of various models using 50 dimensional WE features. A:Disease name recognition, B: Disease classification task}
\label{tab:tab4}
\end{minipage}
\end{table*}
\subsection*{Evaluation of different models with WE and WE+CE }
Next we investigated the results obtained by the various models using only $50$ dim word embedding features. The first part of table \ref{tab:tab4} shows the results obtained by different RNNs and the window based neural network (NN). In this case RNN models are giving better results than the NN model for both the tasks. In particular performance of Bi-LSTM models are best than others in both the tasks. We observe that for the task A, RNN models obtained $1.2\%$ to $3\%$ improvement in F1-score than the baseline NN performance. Similarly $2.55\%$ to $4\%$ improvement in F1-score are observed for the task B, with Bi-LSTM model obtaining more than $4\%$ improvement.

In second part of this table we compare the results obtained by various models using the features set obtained by combining the two feature sets. If we look at performance of individual model using three different set of features, model using only word embedding features seems to give consistently best performance. Among all models, Bi-LSTM using word embedding features obtained best F1-scores of $79.13\%$ and $63.16\%$ for the tasks A and B respectively.

\subsection*{Importance of tuning pre-trained word vectors}
We further empirically evaluate the importance of updating of word vectors while training. For this, we performed another set of experiments, where pre-trained word
vectors are not updated while training. Results obtained on the validation dataset of the Task A are shown in the Table~\ref{tab:tab5}. One can observe that performance of all models have deteriorated. Next, instead of using pre-trained word vectors, we initialized each word with zero vector but kept updating them while training. Although performance
(Table~\ref{tab:tab6}) deteriorated (compare to Table~\ref{tab:tab4}) but not as much as in table~\ref{tab:tab5}. This observation highlights the importance of tuning word vectors for a specific task during training.

\begin{table}[ht]
\centering
\begin{tabular}{|c|c|c|c|c|c|}
\hline
\textit{\textbf{Model}} & \textit{\textbf{P}} & \textit{\textbf{R}} & \textit{\textbf{F}} \\ \hline 
NN+WE	   & 74.02 & 67.86 & 70.81	\\ \hline
Bi-RNN+WE  & 72.17 & 64.40 & 68.06\\ \hline
Bi-GRU+WE  & 77.06 & 70.55 & 73.66 	\\ \hline
Bi-LSTM+WE & 77.32 & 73.75 & 75.49 	\\ \hline
 \end{tabular}
\caption{Performance of different models with {\it 50 dim embedded vectors} in {\bf Task A} {\it validation set} {\bf when word vectors are not getting updated while training}}
\label{tab:tab5}
\end{table}
\begin{table}[ht]
\centering
\begin{tabular}{|c|c|c|c|c|c|}
\hline
\textit{\textbf{Model}} & \textit{\textbf{P}} & \textit{\textbf{R}} & \textit{\textbf{F}} \\ \hline 
NN+RV  		& 81.64 & 74.01 & 77.64 \\ \hline
Bi-RNN+RV 	& 82.32 & 72.73 & 77.2 	\\ \hline
Bi-GRU+RV 	& 82.48 & 74.14 & 78.08 	 \\ \hline
Bi-LSTM+RV	& 83.41 & 72.73 & 77.70 	\\ \hline
 \end{tabular}
\caption{Results of different models with {\it 50 dim random vectors} in {\bf Task A} {\it validation set}}
\label{tab:tab6}
\end{table}
 
\subsection*{Comparison with State-of-art}
At the end we are comparing our results with state-of-the art results reported in \cite{Dogan12} on this dataset using BANNER \cite{LeamanG08} in table \ref{tab:tab3}. BANNER is a CRF based bio entity recognition model, which uses general linguistic, orthographic, syntactic dependency features. Although the result reported in \cite{Dogan12} (F1-score = $81.8$) is better than that of our RNN models but it should be noted that competitive result (F1-score = $79.13\%$) is obtained by the proposed Bi-LSTM model which does not depend on any feature engineering or domain-specific resources and is using only word embedding features trained in unsupervised manner on a huge corpus. 
 
For the task B, we did not find any paper except \cite{gang2012}.
\newcite{gang2012} used 
linear soft margin support vector (SVM) machine with a number of hand designed features including dictionary based features. The best performing proposed model shows more than $37\%$ improvement in F1-score (benchmark: $46\%$ vs Bi-LSTM+WE: $63.16\%$).

\begin{table*}[t]
\begin{minipage}{\textwidth}
\centering
\begin{tabular} 
{|p{0.04\linewidth}|p{0.32\linewidth}|p{0.05\linewidth}|p{0.05\linewidth}|p{0.05\linewidth}|p{0.05\linewidth}|p{0.05\linewidth}|p{0.05\linewidth}|}\hline
\multirow{2}{*}{\textbf{Task}} & \multirow{2}{*}{\textbf{Model}} & \multicolumn{3}{c|}{\textbf{Validation Set}} & \multicolumn{3}{c|}{\textbf{Test Set}} \\ \cline{3-8} 
& & \textbf{P} & \textbf{R} & \textbf{F} & \textbf{P} & \textbf{R} & \textbf{F} \\ \hline 
\hline

\multirow{2}{*}{\textbf{A}} & Bi-LSTM+WE& 85.13 &77.72 & 81.26 & 84.87 & 74.11 & 79.13 \\ \cline{2-8}

& BANNER \cite{Dogan12} & -  & - & {\bf 81.9} & -  & - &  {\bf 81.8} \\ \hline
\hline
\multirow{2}{*}{\textbf{B}} & Bi-LSTM+WE &  67.48 & 58.01 & {\bf 62.39} & 68.97 & 58.25 & {\bf 63.16} \\ \cline{2-8} 

& SM-SVM\cite{gang2012} & - & - & - & 66.1  & 35.2 & 46.0 \\ \hline
\end{tabular}
\caption{Comparisons of our best model results and state-of-art results.  SM-SVM :Soft Margin Support Vector Machine}
\label{tab:tab3}
\end{minipage}
\end{table*}

\section{Failure Analysis}
To see where exactly our models failed to recognize diseases, we analyzed the results carefully. We found that significant proportion of errors are coming due to use of acronyms of diseases and use of disease form which is rarely appearing in our corpus. Examples of few such cases are  {\it ``CD'', ``HNPCC'',``SCA1''}. We observe that this error is occurring because we do not have exact word embedding for these words. Most of the acronyms in the disease corpus were mapped to rare-word embedding\footnote{we obtained pre-trained word-embedding features from \cite{muneeb15} and in their pre-processing strategy, all words of frequency less than $50$ were mapped to rare-word.}. Another major proportion of errors in our results were due to difficulty in recognizing nested forms of disease names. For example, in all of the following cases: {\it ``hereditary forms of 'ovarian cancer'" , ``inherited `breast cancer'", ``male and female `breast cancer'"}, part of phrase such as {\it ovarian cancer} in {\it hereditary forms of ovarian cancer}, {\it breast cancer} in {\it inherited breast cancer} and {\it male and female breast cancer} are disease names and our models are detecting this very well. However, according to annotation scheme if any disease is part of nested disease name, annotators considered whole phrase as a single disease. So even our model is able to detect part of the disease accurately but due to the exact matching scheme, this will be false positive for us. 
 
\section{Related Research}
\label{sec:rel_work}

In biomedical domain, named entity recognition has attracted much attention for identification of entities such as genes and proteins \cite{abner05,LeamanG08,leaman09} but not as much for disease name recognition. Notable works, such as of Chowdhury and Lavelli (2010), are mainly conditional random field (CRF) based models using lots of manually designed template features. These include linguistic, orthographic, contextual and dictionary based features. However, they have evaluated their model on the AZDC dataset which is small compared to the NCBI dataset, which we have considered in this study. \newcite{nikfarjam2015} have proposed a CRF based sequence tagging model, where cluster id of embedded word as an extra feature with manually engineered features is used for adverse drug reaction recognition in tweets.


Recently deep neural network models with minimal dependency on feature engineering have been used in few studies in NLP including NER tasks \cite{collobert11a,collobert08}. dos Santos et al. \shortcite{dos15a} used deep neural network based model such as window based network to recognize named entity in Portuguese and Spanish texts. In this work, they exploit the power of CNN to get morphological and shape features of words in character level word embedding, and used it as feature with concatenation of word embedding. Their results indicate that CNN are able to preserve morphological and shape features through character level word embedding. Our models are quite similar to this model but we used different variety of RNN in place of window based neural network. 

\newcite{LabeauLA15} used Bi-RNN with character level word embedding only as a feature for PoS tagging in German text. Their results also show that with only character level word embedding we can get state-of-art results in PoS tagging in German text. Our model used word embedding as well as character level word embedding together as features and also we have tried more sophisticated RNN models such as LSTM and GRU in bi-directional structure. More recent work of \newcite{HuangXY15} used LSTM and CRF in variety of combination such as only LSTM, LSTM with CRF and Bi-LSTM with CRF for PoS tagging, chunking and NER tasks in general texts. Their results shows that Bi-LSTM with CRF gave best results in all these tasks. These two works have used either Bi-RNN with character embedding features or Bi-LSTM with word embedding features in general or news wire texts, whereas in this work we compare the performance of three different types of RNNs: Bi-RNN, Bi-GRU and Bi-LSTM with both word embedding and character embedding features in biomedical text for disease name recognition. 

\section{Conclusions}
In this work, we used three different variants of bidirectional RNN models with word embedding features for the first time for disease name and class recognition tasks. Bidirectional RNN models are used to capture both forward and backward long term dependencies among words within a sentence. We have shown that these models are able to obtain quite competitive results compared to the benchmark result on the disease name recognition task. Further our results have shown a significantly improved results on the relatively harder task of disease classification which has not been studied much. All our results were obtained without putting any effort on feature engineering or requiring domain-specific knowledge. Our results also indicate that RNN based models perform  better than window based neural network model for the two tasks. This could be due to the implicit ability of RNN models to capture variable range dependencies of words compared to explicit dependency on context window size of window based neural network models.

\section*{Acknowledgments}
We acknowledge the use of computing resources made available from the Board of Research in Nuclear Science (BRNS), Dept of Atomic Energy (DAE)
Govt. of India sponsered project (No.2013/13/8-BRNS/10026) by Dr Aryabartta Sahu at Department of Computer Science and Engineering, IIT Guwahati. 


\bibliography{acl2016}
\bibliographystyle{acl2016}

 
 

\end{document}